\documentclass[final]{cvpr}

\usepackage{times}
\usepackage{epsfig}
\usepackage{graphicx}
\usepackage{amsmath}
\usepackage{amssymb}

\usepackage{multirow}
\usepackage[table,xcdraw]{xcolor}
\usepackage{url}            
\usepackage{booktabs}       
\usepackage{amsfonts}       
\usepackage{nicefrac}       
\usepackage{microtype}      
\usepackage{setspace}
\usepackage[ruled,vlined]{algorithm2e}    
\SetKwInput{KwInput}{Input}
\SetKwInput{KwOutput}{Output}
\usepackage{nameref}
\usepackage{pifont}
\DeclareMathOperator*{\argmax}{arg\,max}

\newcommand{\lam}{\lambda}

\usepackage{stfloats}
\usepackage{pdflscape}
\usepackage{arydshln}

\usepackage[pagebackref=true,breaklinks=true,colorlinks,linkcolor=blue,bookmarks=false]{hyperref}

\def\cvprPaperID{10370} 
\def\confYear{CVPR 2021}

\begin{document}

\title{PixMatch: Unsupervised Domain Adaptation via Pixelwise Consistency Training}

\author{Luke Melas-Kyriazi\\
Harvard University\\
Oxford University\\
Cambridge, MA\\
{\tt\small lukemk@robots.ox.ac.uk}
\and
Arjun K. Manrai \\
Harvard University \\
Boston Children's Hospital \\
Boston, MA \\
{\tt\small Arjun\_Manrai@hms.harvard.edu}
}

\maketitle

\begin{abstract}
Unsupervised domain adaptation is a promising technique for semantic segmentation and other computer vision tasks for which large-scale data annotation is costly and time-consuming. In semantic segmentation, it is attractive to train models on annotated images from a simulated (source) domain and deploy them on real (target) domains. 
In this work, we present a novel framework for unsupervised domain adaptation based on the notion of target-domain consistency training. 
Intuitively, our work is based on the idea that in order to perform well on the target domain, a model’s output should be consistent with respect to small perturbations of inputs in the target domain. Specifically, we introduce a new loss term to enforce pixelwise consistency between the model's predictions on a target image and a perturbed version of the same image.  
In comparison to popular adversarial adaptation methods, our approach is simpler, easier to implement, and more memory-efficient during training. Experiments and extensive ablation studies demonstrate that our simple approach achieves remarkably strong results on two challenging synthetic-to-real benchmarks, GTA5-to-Cityscapes and SYNTHIA-to-Cityscapes. 
\end{abstract}
\section{Introduction}

Deep neural network approaches for semantic image segmentation have shown widespread success in the past decade, but they remain reliant on large datasets with pixel-level annotations. Data labeling for semantic segmentation is notoriously laborious and expensive, especially in domains where experts are required (e.g. medical image segmentation). Even for annotations that can be performed by non-experts like parsing an urban scene into familiar objects, as in the Cityscapes dataset \cite{cityscapes}, it takes an estimated 90 minutes to annotate a single image \cite{statistic}. 

The need to build generalizable models with limited data has motivated work on unsupervised domain adaptation (UDA) approaches for semantic segmentation \cite{cbst,adaptsegnet,advent,maxsquare,cycada,chen2018road}, where annotated images from a simulated (source) domain, which are plentiful, are used in conjunction with unlabeled images from a real (target) domain. The simulated source domain in this ``synthetic to real'' translation task can be creative, such as the video game Grand Theft Auto V in the GTA5-to-Cityscapes benchmark \cite{gta} and the simulation platform SYNTHIA as in the SYNTHIA-to-Cityscapes benchmark \cite{synthia}. 

The literature on UDA for semantic segmentation is dominated by adversarial methods, which aim to learn domain-invariant representations across multiple domains by introducing adversarial losses \cite{cycada}. These methods have shown strong performance, but due to the instability of their adversarial losses, they are well-known to be highly sensitive to hyperparameters and difficult to train \cite{gans_hard_one,gans_hard_two,wgan}. 

Recently, a new line of work on UDA for semantic segmentation has emerged around self-training \cite{cbst,advent,maxsquare}. These methods add loss terms to the training objective that encourage the segmentation model to make more confident predictions on the target domain (for example, by encouraging low-entropy predictions) \cite{advent,maxsquare,cbst}. 

This paper begins with the observation that we do not simply desire a model that makes confident predictions in target domain, rather we desire a model that makes \textit{consistent} predictions in the target domain. That is, we desire a model for which small perturbations of inputs in the target domain lead to small, consistent changes in the output segmentation. If a model's predictions are always confident, but they are not stable with respect to small perturbations of target images, the model is likely to be poorly-adapted to the target domain. Conversely, if a model behaves smoothly with respect to perturbations of images in the target domain, the model is likely to be better-adapted to that domain. 

We propose a consistency training-based framework to directly enforce this notion of smoothness in the target domain. Our method, denoted PixMatch, adds a loss term that encourages the segmentation model's predictions on a target domain image and a perturbed version of the same image to be pixelwise consistent. 

We experiment with four different perturbation functions, two of which are inspired by work in semi/self-supervised learning (Data Augmentations; CutMix) and two of which are inspired by work in domain adaptation (Style Transfer; Fourier Transform). 

Surprisingly, we find that our baseline model, which uses heavy data augmentation as its perturbation function, performs best. This simple baseline delivers extremely strong results on GTA5-to-Cityscapes \cite{cityscapes} and SYNTHIA-to-Cityscapes \cite{synthia}. Using only a source (supervised) loss and a target consistency loss, it outperforms complex prior methods that used combinations of source, adversarial, and self-training losses. 

Compared to existing adversarial approaches, PixMatch is easier to implement, more stable during training, and less memory-intensive to train. It introduces only one hyperparameter, which controls the relative weighting of the source (supervised) loss and the target consistency loss. Moreover, the simplicity of PixMatch means that it it may be easily integrated into existing UDA methods and pipelines; combining PixMatch with self-training yields the state-of-the-art results. 

The main contributions of this paper are:

\begin{itemize}
\item We introduce a novel consistency-based framework for unsupervised domain adaptation that encourages pixelwise consistency between a model's predictions on a target image and a perturbed version of the same image. 
\item We investigate multiple perturbation functions, finding that a simple baseline using data augmentation performs extremely well. 
\item We perform extensive ablation studies on our baseline method in order to better understand its strong performance, showing that the setting of domain adaptation differs markedly from that of semi-supervised image classification. 
\item We find that our model may be easily combined with self-training for further performance improvements. Doing so, we achieve a new state-of-the-art on the challenging GTA5-to-Cityscapes benchmark, using a much simpler training approach than recent adversarial methods.
\end{itemize}

\section{Background}


The task of semantic image segmentation is to assign to every pixel in an image one of $C$ semantic class labels. In the setting of unsupervised domain adaptation (UDA) for semantic segmentation, we consider a labeled source domain $S$ and an unlabeled target domain $T$ with the same set of semantic classes. Often, the source domain $S$ is composed of simulated data (e.g. GTA V) and the target $T$ is composed of real data (e.g. Cityscapes), thus forming a ``synthetic-to-real'' adaptation task. 


\subsection{Related work}
\subsubsection{Unsupervised Domain Adaptation for Semantic Segmentation}

Recent work on UDA for semantic segmentation has been dominated by deep neural network methods. 

Most of the literature is concerned with adversarial training \cite{cycada,adaptsegnet,advent,chen2018road}. Adversarial training aims to minimize the discrepancy between source and target feature distributions by introducing a discriminator network alongside the main segmentation network. The discriminator is trained to predict an input image's domain from the segmentation network's intermediate feature maps, while the segmentation network is trained to fool the discriminator (and produce good segmentations on the source domain). Adversarial training generally produces strong performance but suffers from instability during training, is computationally expensive, and is highly sensitive to changes in hyperparameters. 

During the past two years, a new line of work has emerged around self-training. Zou et al. \cite{cbst} were the first to apply pseudolabeling to UDA; they alternatively generate pseudolabels on target data and re-train the model with these labels. They also use different pseudolabel thresholds for each class (``class-balanced self-training'') to prevent the loss from being dominated by easy classes. 

Vu et al. \cite{advent} propose two entropy minimization approaches, one that operates directly (MinEnt) and one that operates adversarially (AdvEnt). Chen et al. \cite{maxsquare} builds off AdvEnt \cite{advent} by noting that the entropy function $H(\cdot)$ biases the loss toward well-classified pixels in an image rather than more challenging pixels. To address this, Chen et al. \cite{maxsquare} substitute a linear function in place of the entropy function $H$. These methods demonstrate competitive performance with adversarial approaches on GTA5-to-Cityscapes and SYNTHIA-to-Cityscapes. 

\begin{figure}
  \includegraphics[width=0.5\textwidth]{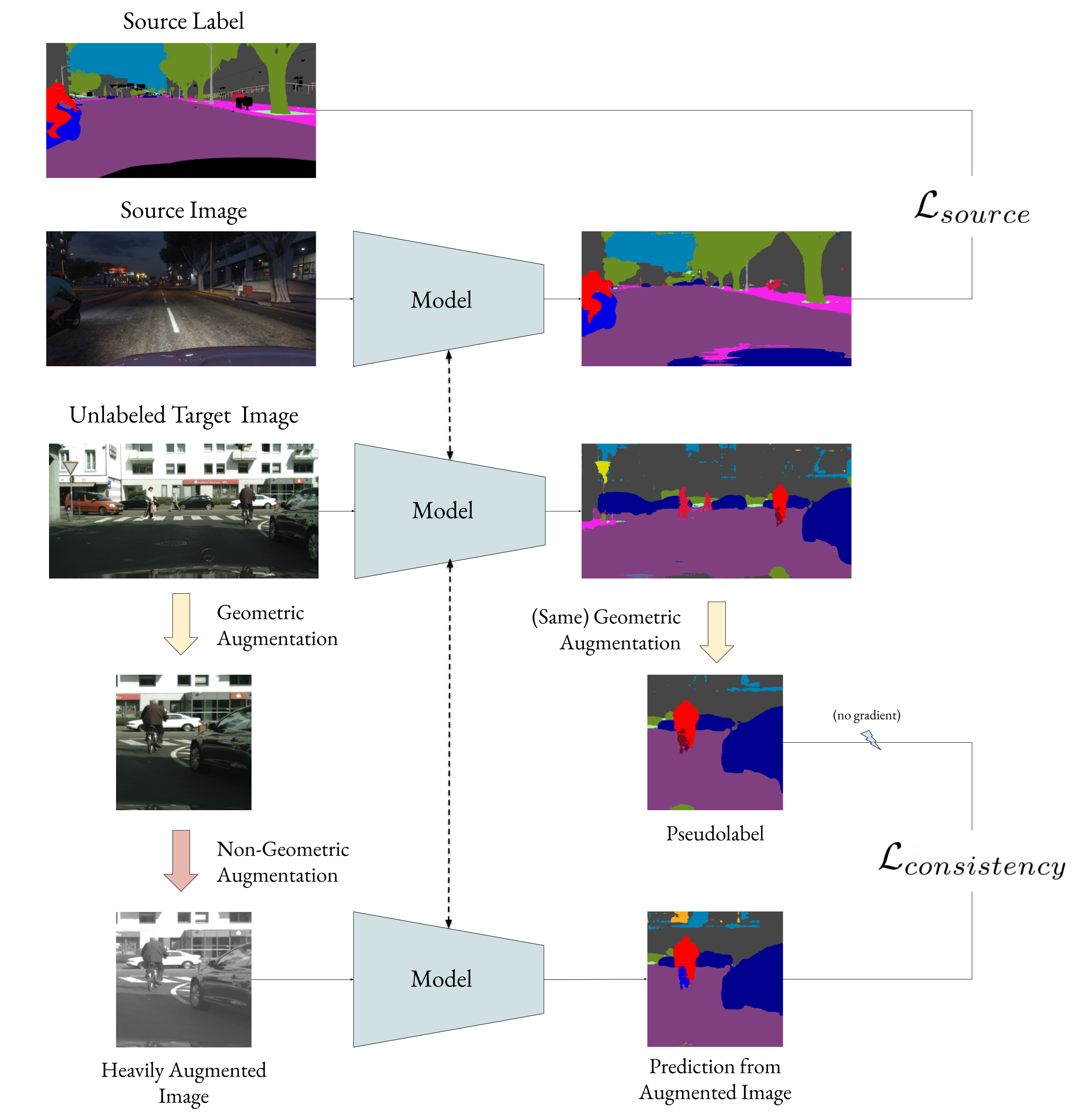}
  \caption{Our proposed pixelwise consistency training approach.}
  \label{fig:diagram}
\end{figure}

Some recent work in UDA for semantic segmentation has considered the broad notion of consistency, but they do so in a different manner from this work. Most notably, a significant body of research leverages cycle-consistency to translate images in the source domain to the target domain for domain adaptation \cite{cycada,chen2019crdoco,fu2019geometry,Li_2019_CVPR,zhao2019multi,murez2018image}. \cite{cycada} employs cycle-consistent GANs for domain adaptation at both the pixel-level and feature-level. \cite{murez2018image} trains a GAN-based image-to-image translation network and a semantic segmentation network with the same backbone, and \cite{Li_2019_CVPR} extends the idea of cross-domain cycle-consistency to the multi-source setting. \cite{li2019bidirectional} trains an image-to-image translation model and a semantic segmentation model in alternating stages, while the very recent work \cite{kang2020pixel} considers a contrastive cycle-consistency loss on the level of pixels. Finally, \cite{yang2020fda} performs image-to-image translation for UDA in frequency space rather than pixel space using a Fourier transform.

Beyond cycle-consistency, \cite{chen2019crdoco} enforces cross-domain consistent predictions in the context of image-to-image translation for UDA for using two image-level adversarial losses and two feature-level adversarial losses. 
\cite{fu2019geometry} enforces geometric consistency in the context of image-to-image translation, \cite{kim2020learning} aims to learn texture-invariant representations, and \cite{iqbal2020mlsl} aims to learn spatially-consistent features by segmenting images on the level of patches. 

Our formulation of consistency training differs from all of these above approaches in that it directly applies a general consistency loss to the semantic segmentation network in the target domain, ensuring that the output is consistent with respect to perturbations. Our framework is flexible, supporting perturbation functions inspired by work from both the semi/self-supervised learning community (below) and the domain adaptation community. Whereas most of the previous approaches involve adversarial losses, auxiliary networks, or complex multi-stage training pipelines, our framework involves only a single segmentation network with a single non-adversarial consistency loss (in addition to the standard source cross-entropy loss). Additionally, our framework is flexible and may be used alongside any of these prior approaches.

\subsubsection{Consistency Training}

Consistency training was first proposed for semi-supervised learning by Blum and Mitchell \cite{cotraining} and enjoyed widespread use throughout the 2000s under the name ``co-training''. This co-training framework was slightly different than the approach we have today; Blum and Mitchell \cite{cotraining} trained multiple models on different views of the same set of input examples and used these models to produce pseudolabels for one another. Modern consistency training \cite{sajjadi2016regularization} produces multiple views of the same input using a stochastic data augmentation function, and uses them to train a single model. 

Recently, consistency training has formed the base of multiple large advances in semi-supervised learning for image classification. Xie et al. \cite{xie2019unsupervised} use a model to generate (soft) artificial labels and enforce strongly-augmented examples, showing strong performance on CIFAR-10 with as few as 250 examples. In MixMatch \cite{berthelot2019mixmatch}, Berthelot et al. mix labeled and unlabeled data while performing entropy minimization on augmented unlabeled data. In their follow-up ReMixMatch \cite{berthelot2019remixmatch}, they further improve the method's sample-efficiency by optimizing the data augmentation function during training. Most recently, in FixMatch \cite{fixmatch}, Sohn et al. replace entropy minimization with pseudolabeling followed by filtering for high-confidence examples. 

We denote our method ''PixMatch`` as a reference to this line of work, but we emphasize that it is designed for the different setting of unsupervised domain adaptation, not semi-supervised image classification. In this paper we are interested in answering the question of what forms of consistency losses are effective \textit{for UDA specifically}. 

\begin{algorithm}[t]
  \setstretch{1.15}
  \fontsize{8}{9}\selectfont
  \SetAlgoLined
  \DontPrintSemicolon
  \KwInput{Input image $x \in \mathbb{R}^{H \times W \times 3}$}
  \KwResult{Target consistency loss} 
    \hskip0.2em $y_{pseudo}$ $=$ model$(x)$ \textit{\# without gradient}  \;
    \hskip0.2em $x_{pert}$, $y_{pseudo\_pert}$ $=$ perturbation$_{S}(x, y_{pseudo}$)\;
    \hskip0.2em $y_{hat}$ $=$ model$(x_{pert})$\;
    \hskip0.2em loss $=$ cross\_entropy$(y_{pseudo\_pert}, y_{hat})$\;
    \hskip0.2em \textbf{return }loss
  \caption{\small Target Consistency Loss}
  \end{algorithm}

\subsubsection{Contrastive Learning and Data Augmentation}

Consistency training is closely related to contrastive learning, which has recently led to remarkable progress in unsupervised representation learning \cite{moco,simclr}. 
Of particular relevance for this paper is SimCLR \cite{simclr}; our baseline PixMatch model uses the data augmentations proposed in SimCLR as its perturbation function for consistency training. 
In related work, \cite{french2018selfensembling} uses a student-teacher architecture with augmentations for image classification. 



\section{PixMatch}

\subsection{Overview}

PixMatch employs consistency training and pseudolabeling to enforce consistency on the target domain. Its loss function is composed of two cross-entropy loss terms. The first of these is the standard supervised cross-entropy loss on the source domain:
\begin{equation}
  \mathcal{L}_S = -\frac{1}{n_S} \sum_{s \in S} \sum_{i=1}^{H \cdot W} H(y_s^{(i)}, p^{(i)}(y | x_s))
\end{equation}
where $p^{(i)}$ is the output probability distribution at pixel $i$ for source input $x_s$, $y$ is the ground truth semantic map, and $n_S$ is the number of images in the source dataset $S$.

The second of these losses is a consistency loss on the target domain. To calculate this loss, we first run a target image $x_t$ through the model to obtain the pseudolabel $\hat{y_t} = \argmax(q_t) = \argmax(p_t(y | x_t))$. We then perturb the (image, pseudolabel) pair using a perturbation function to yield the pair $(x_{pert}, \hat{y_{pert}})$. Our consistency loss function is then 
\begin{equation}
  \mathcal{L}_T = -\frac{1}{n_T} \sum_{t \in T} \sum_{i=1}^{H \cdot W} H(\hat{y_{pert}}^{(i)}, p^{(i)}(y | x_{pert}))
\end{equation}
and the final loss is $\mathcal{L} = \mathcal{L}_S + \lambda_T \mathcal{L}_T$, where $\lambda_T$ is a hyperparameter that controls the relative weighting of the source (supervised) loss and the target consistency loss.

\subsection{Perturbation Functions}

We experiment with four perturbation functions, two inspired by the self-supervised learning community and two inspired by the domain adaptation community. 

\paragraph{Data Augmentation Consistency} Our baseline PixMatch model uses heavy data augmentation as the perturbation function (\autoref{fig:sup_algos}). It is important to note that, unlike in image classification, when we augment an input target image using a geometric transform (e.g. cropping), we also have to perform the corresponding augmentation on the pseudolabel produced by the model. 

\paragraph{CutMix Consistency} Our second model uses the recently-proposed CutMix \cite{yun2019cutmix} regularization method as a perturbation function. Specifically, we cut-and-paste a rectangular region from a source image onto the target image and enforce that the predictions on this mixed image are consistent with the corresponding mixed labels. Note that this perturbation function takes as input both the source and target image, rather than just the target image. We use CutMix rather than other regularization methods because, in comparison to MixMatch \cite{berthelot2019mixmatch} and others, \cite{yun2019cutmix} argues that CutMix promotes the learning of more localizable features. 
This perturbation function is closely related to the approach taken in \cite{cutmix_for_ssss} for semi-supervised semantic segmentation.

\paragraph{Style Consistency} Our third perturbation function is inspired by the line of work in domain adaptation that tries to transfer the style of the target images to the source prior to or during training \cite{cycada,adaptsegnet,chen2019crdoco,murez2018image}. Here, we transfer the style of the source to the target and enforce consistency between the model's predictions on the original target image and the source-stylized target image. To perform the target-to-source stylization, we use the zero-shot style transfer model from \cite{sheng2018avatar} (AvatarNet). 

\paragraph{Fourier Consistency} Our fourth perturbation function, inspired by but different from \cite{yang2020fda}, modifies the target image in frequency space using a Fourier transform. \cite{yang2020fda} replaces the low-frequency part of source images with that of target images, creating source images with low-frequency target styles; they train using a cross-entropy loss between these modified source images and the source labels, alongside entropy minimization on the target images. Differently, we replace the low-frequency part of the target images with that of the source, creating target images with low-frequency source styles; we train with a consistency loss between the original images and these Fourier-stylized target images, alongside the standard cross-entropy loss on the source. 

\subsection{Relationship between PixMatch and Self-Training}

As a general framework, PixMatch is closely related to self-training methods. In particular, if the perturbation function is the identity function, then the pseudolabels produced during consistency training are the standard pseudolabels from self-training. It is also possible to perform PixMatch in a soft manner, analogous to the way in which entropy minimization is a soft variant of pseudolabeling. In our ablation studies on our baseline model, we find that a hard approach gives better empirical results. 

However, our framework is different from other self-training methods in that it enables domain-adaptation-specific self-training; the perturbation function may incorporate information from the source data. For example, our Fourier-based consistency function uses source-domain Fourier frequencies to perturb the target images. By contrast, other pure self-training approaches \cite{cbst,maxsquare,advent} use only target-domain information. 

\section{Experiments}
\paragraph{Datasets}

We evaluate our method on the two standard large-scale UDA segmentation benchmarks, GTA5-to-Cityscapes and SYNTHIA-to-Cityscapes. For evaluation, we measure per-class and mean Intersection-over-Union (IoU). As is standard, we evaluate using all 19 classes for GTA5-to-Cityscapes, and using both 16 and 13 classes for SYNTHIA-to-Cityscapes (because SYNTHIA shares only 16 classes with Cityscapes). 

\paragraph{Model}

For purposes of comparison with previous methods, we use the DeepLab-v2 segmentation model \cite{deeplab} with a ResNet-101 \cite{resnet} backbone. We train using SGD with learning rate $1.0 \cdot 10^{-4}$, momentum $0.9$, weight decay $5 \cdot 10^{-4}$, and a polynomial learning rate decay: $lr = lr * (1 - \tfrac{iter}{max\_iter})^{0.9}$. All experiments are conducted on a single NVIDIA RTX 2080TI GPU with 11GB of VRAM. The exact parameters used for each perturbation function are included in the appendix. 


\begin{table*}[h]
\fontsize{8}{9}\selectfont
\setlength{\tabcolsep}{0.30em}
\def\arraystretch{1.15}
\arrayrulecolor{gray}
\begin{center}
\addtolength{\leftskip} {-0.5cm} 
\textbf{GTA5-to-Cityscapes}

\vspace{1mm}
\begin{tabular}{lll|c|ccccccccccccccccccc|c}
                    & Year & Method & Backbone & \rotatebox{90}{road} & \rotatebox{90}{sdwk} & \rotatebox{90}{bld} & \rotatebox{90}{wall} & \rotatebox{90}{fnc} & \rotatebox{90}{pole} & \rotatebox{90}{lght} & \rotatebox{90}{sign} & \rotatebox{90}{veg.} & \rotatebox{90}{trrn.} & \rotatebox{90}{sky} & \rotatebox{90}{pers} & \rotatebox{90}{rdr} & \rotatebox{90}{car} & \rotatebox{90}{trck} & \rotatebox{90}{bus} & \rotatebox{90}{trn} & \rotatebox{90}{mtr} & \rotatebox{90}{bike} & mIoU \\ \toprule    
\multicolumn{24}{c}{No Adaptation} \\ \midrule 
    &   & Source only	                        & RN-101 &			71.4 & 15.3 & 74.0 & 21.1 & 14.4 & 22.8 & 33.9 & 18.6 & 80.7 & 20.9 & 68.5 & 56.6 & 27.1 & 67.4 & 32.8 &  5.6 & 7.7 & 28.4 & 33.8 & 36.9 \\ 
\midrule
\multicolumn{24}{c}{Adversarial Methods} \\ 
\midrule 
\cite{adaptsegnet}         & \textit{2018} \hspace{2mm} &  AdaptSegNet         &  RN-101 &		      86.5 &  25.9                                                                                           &  79.8                      &  22.1                &  20.0 &  23.6 &  33.1 &  21.8 &  81.8 &  25.9 &  75.9 &  57.3 &  26.2 &  76.3 &  29.8 &  32.1 &  7.2  &  29.5 &  32.5 &  41.4 \\ 
\cite{adaptsegnet}         & \textit{2018} \hspace{2mm} &  AdaptSegNet-LS    &  RN-101 &		91.4       &  48.4                                                                                           &  81.2                      &  27.4                &  21.2 &  31.2 &  35.3 &  16.1 &  84.1 &  32.5 &  78.2 &  57.7 &  28.2 &  85.9 &  33.8 &  43.5 &  0.2  &  23.9 &  16.9 &  44.1 \\
\cite{advent}              & \textit{2019} \hspace{2mm} &  AdvEnt              &  RN-101               &  89.9                                                                                           &  36.5                      &  81.6                &  29.2 &  25.2 &  28.5 &  32.3 &  22.4 &  83.9 &  34.0 &  77.1 &  57.4 &  27.9 &  83.7 &  29.4 &  39.1 &  1.5  &  28.4 &  23.3    &  43.8 \\
\cite{advent}              & \textit{2019} \hspace{2mm} &  AdvEnt+MinEnt       &  RN-101               &  89.4                                                                                           &  33.1                      &  81.0                &  26.6 &  26.8 &  27.2 &  33.5 &  24.7 &  83.9 &  36.7 &  78.8 &  58.7 &  30.5 &  84.8 &  38.5 &  44.5 &  1.7  &  31.6 &  32.4    &  45.5 \\
\cite{discpatch}           & \textit{2019} \hspace{2mm} &  Patch-Disc          &  RN-101               &  92.3                                                                                           &  51.9                      &  82.1                &  29.2 &  25.1 &  24.5 &  33.8 &  33.0 &  82.4 &  32.8 &  82.2 &  58.6 &  27.2 &  84.3 &  33.4 &  46.3 &  2.2  &  29.5 &  32.3    &  46.5 \\
\cite{zhang2019category}   & \textit{2019} \hspace{2mm} &  CAG                 &  RN-101               &  90.4& 51.6& 83.8& 34.2& 27.8& 38.4& 25.3& 48.4& 85.4& 38.2& 78.1& 58.6& 34.6& 84.7& 21.9& 42.7 &  41.1& 29.3& 37.2& 50.2 \\ 
\cite{Wang2020ClassesMA}   & \textit{2020} \hspace{2mm} &  FADA                &  RN-101               &  92.5& 47.5& 85.1& 37.6& 32.8& 33.4& 33.8& 18.4& 85.3& 37.7& 83.5& 63.2& 39.7& 87.5& 32.9& 47.8 &  1.6                       &  34.9& 39.5& 49.2 \\ 
\cite{Wang2020ClassesMA}   & \textit{2020} \hspace{2mm} &  FADA-MST            &  RN-101               &  91.0& 50.6& 86.0& 43.4& 29.8& 36.8& 43.4& 25.0& 86.8& 38.3& 87.4& 64.0& 38.0& 85.2& 31.6& 46.1 &  6.5                       &  25.4& 37.1& 50.1 \\
\cite{huang2020contextual} & \textit{2020} \hspace{2mm} &  CrCDA               &  RN-101               &  92.4& 55.3& 82.3& 31.2& 29.1& 32.5& 33.2& 35.6& 83.5& 34.8& 84.2& 58.9& 32.2& 84.7& 40.6& 46.1 &  2.1                       &  31.1& 32.7& 48.6 \\
\midrule
\multicolumn{24}{c}{Image-to-Image / Style Transfer Methods} \\ 
\midrule 
\cite{li2019bidirectional} & \textit{2019} \hspace{2mm} &  BDL (M2-F2)    &  RN-101 &  91.0& 44.7& 84.2& 34.6& 27.6& 30.2& 36.0& 36.0& 85.0& 43.6& 83.0& 58.6& 31.6& 83.3& 35.3& 49.7                                      &  3.3 &  28.8& 35.6& 48.5 \\
\cite{chen2019crdoco}      & \textit{2019} \hspace{2mm} &  CrDoCo         &  DRN-26 &  95.1& 49.2& 86.4& 35.2& 22.1& 36.1& 40.9& 29.1& 85.0& 33.1& 75.8& 67.3& 26.8& 88.9& 23.4& 19.3                                      &  4.3 &  25.3& 13.5& 45.1 \\
\cite{yang2020fda}         & \textit{2020} \hspace{2mm} &  FDA            &  RN-101 &  88.8& 35.4& 80.5& 24.0& 24.9& 31.3& 34.9& 32.0& 82.6& 35.6& 74.4& 59.4& 31.0& 81.7& 29.3& 47.1                                      &  1.2 &  21.1& 32.3& 44.6 \\
\cite{yang2020fda}         & \textit{2020} \hspace{2mm} &  FDA (Ensemble) &  RN-101 &  92.5& 53.3& 82.3& 26.5& 27.6& 36.4& 40.5& 38.8& 82.2& 39.8& 78.0& 62.6& 34.4& 84.9& 34.1& 53.12& 16.8& 27.7& 46.4& \textit{50.4} \\ 
\cite{kim2020learning}     & \textit{2020} \hspace{2mm} &  LTIR           &  RN-101 &  92.9& 55.0& 85.3& 34.2& 31.1& 34.9& 40.7& 34.0& 85.2& 40.1& 87.1& 61.0& 31.1& 82.5& 32.3& 42.9                                      &  0.3 &  36.4& 46.1& 50.2 \\
\midrule 
\multicolumn{24}{c}{Self-Training Methods} \\ 
\midrule 
\cite{cbst}                        & \textit{2018} \hspace{2mm} &  CBST-SP                    &  WRN-38                                                                                         &  88.0                                                                                           &  56.2                      &  77.0 &  27.4 &  22.4 &  40.7 &  47.3 &  40.9 &  82.4 &  21.6 &  60.3 &  50.2 &  20.4 &  83.8 &  35.0 &  51.0 &  15.2 &  20.6 &  37.0 &  46.2 \\
\cite{advent}                      & \textit{2019} \hspace{2mm} &  MinEnt                     &  RN-101                                                                                         &  86.2                                                                                           &  18.6                      &  80.3 &  27.2 &  24.0 &  23.4 &  33.5 &  24.7 &  83.3 &  31.0 &  75.6 &  54.6 &  25.6 &  85.2 &  30.0 &  10.9 &  0.1  &  21.9 &  37.1 &  42.3 \\
\cite{maxsquare}                   & \textit{2019} \hspace{2mm} &  MaxSquare (MS)             &  RN-101                                                                                         &  88.1                                                                                           &  27.7                      &  80.8 &  28.7 &  19.8 &  24.9 &  34.0 &  17.8 &  83.6 &  34.7 &  76.0 &  58.6 &  28.6 &  84.1 &  37.8 &  43.1 &  7.2  &  32.2 &  34.2 &  44.3 \\
\cite{maxsquare}                   & \textit{2019} \hspace{2mm} &  MS+IW+Multi                &  RN-101                                                                                         &  89.4                                                                                           &  43.0                      &  82.1 &  30.5 &  21.3 &  30.3 &  34.7 &  24.0 &  85.3 &  39.4 &  78.2 &  63.0 &  22.9 &  84.6 &  36.4 &  43.0 &  5.5  &  34.7 &  33.5 &  46.4 \\
\cite{zou2019confidence}           & \textit{2019} \hspace{2mm} &  CRST (MRENT)               &  RN-101                                                                                         &  91.8& 53.4& 80.6& 32.6& 20.8& 34.3& 29.7& 21.0& 84.0& 34.1& 80.6& 53.9& 24.6& 82.8& 30.8& 34.9 &  16.6& 26.4& 42.6& 46.1 \\
\cite{zou2019confidence}           & \textit{2019} \hspace{2mm} &  CRST (MRKLD)               &  RN-101                                                                                         &  91.0& 55.4& 80.0& 33.7& 21.4& 37.3& 32.9& 24.5& 85.0& 34.1& 80.8& 57.7& 24.6& 84.1& 27.8& 30.1 &  26.9& 26.0& 42.3& 47.1 \\
\cite{zou2019confidence}           & \textit{2019} \hspace{2mm} &  CRST (LRENT)               &  RN-101                                                                                         &  91.8& 53.5& 80.5& 32.7& 21.0& 34.0& 29.0& 20.3& 83.9& 34.2& 80.9& 53.1& 23.9& 82.7& 30.2& 35.6 &  16.3& 25.9& 42.8& 45.9 \\ 
\cite{lian2019constructing}        & \textit{2019} \hspace{2mm} &  PyCDA &  RN-101                     &  90.5& 36.3& 84.4& 32.4& 28.7& 34.6& 36.4& 31.5& 86.8& 37.9& 78.5& 62.3& 21.5& 85.6& 27.9& 34.8 &  18.0& 22.9& 49.3& 47.4 \\
\cite{iqbal2020mlsl}               & \textit{2019} \hspace{2mm} &  MLSL (SISC) ST             &  RN-101                                                                                         &  91.0& 49.3& 79.9& 24.4& 27.9& 37.9& 45.1& 45.1& 81.3& 19.0& 61.7& 63.9& 28.0& 86.5& 23.9& 42.3 &  41.9& 33.1& 44.4& 48.7 \\ 
\cite{iqbal2020mlsl}               & \textit{2020} \hspace{2mm} &  MLSL (+PWL) ST& RN-101 &  89.0& 45.2& 78.2& 22.9& 27.3& 37.4& 46.1& 43.8& 82.9& 18.6& 61.2& 60.4& 26.7& 85.4& 35.9& 44.9 &  36.4& 37.2& 49.3& 49.0 \\ 
\midrule
\multicolumn{24}{c}{Other Methods} \\ 
\midrule 
\cite{lee2019sliced}       & \textit{2019} \hspace{2mm} &  SWD     &  RN-101 &  92.0& 46.4& 82.4& 24.8& 24.0& 35.1& 33.4& 34.2& 83.6& 30.4& 80.9& 56.9& 21.9& 82.0& 24.4& 28.7 &  6.1 &  25.0& 33.6& 44.5 \\
\cite{li2020content}       & \textit{2020} \hspace{2mm} &  CCM     &  RN-101 &  93.5& 57.6& 84.6& 39.3& 24.1& 25,2& 35.0& 17.3& 85.0& 40.6& 86.5& 58.7& 28.7& 85.8& 49.0& 56.4 &  5.4 &  31.9& 43.2& 49.9 \\
\cite{kang2020pixel}       & \textit{2020} \hspace{2mm} &  PLCA    &  RN-101 &  84.0& 30.4& 82.4& 35.3& 24.8& 32.2& 36.8& 24.5& 85.5& 37.2& 78.6& 66.9& 32.8& 85.5& 40.4& 48.0 &  8.8 &  29.8& 41.8& 47.7 \\
\cite{pan2020unsupervised} & \textit{2020} \hspace{2mm} &  IntraDA &  RN-101 &  90.6& 37.1& 82.6& 30.1& 19.1& 29.5& 32.4& 20.6& 85.7& 40.5& 79.7& 58.7& 31.1& 86.3& 31.5& 48.3 &  0.0 &  30.2& 35.8& 46.3 \\
\midrule \midrule 
\multicolumn{3}{l|}{\textit{Ours (CutMix)}}             & RN-101 & 90.9 & 45.2 & 81.3 & 26.4 & 19.2 & 21.7 & 33.2  & 18.3 & 84.3  & 41.1 & 78.8 & 61.2 & 21.3 & 87.2 & 43.6 & 50.2 & 5.8 & 29.3 & 21.8 & 45.4 \\
\multicolumn{3}{l|}{\textit{Ours (Style)}}              & RN-101 & 83.7 & 34.5 & 81.5 & 18.9 & 11.0 & 17.6 & 24.3 & 24.0 & 84.7 & 31.5 & 77.9 & 59.8 & 21.2 & 81.3 & 42.5 & 36.2 & 2.51 & 6.30  & 5.23 & 39.2 \\
\multicolumn{3}{l|}{\textit{Ours (Fourier)}}            & RN-101 & 83.0 & 34.4 & 81.1 & 27.9 & 14.4 & 22.3 & 35.1 & 15.7 & 85.3 & 39.4 & 77.3 & 59.1 & 18.2 & 86.6 & 43.3 & 49.5 & 5.46 & 30.2 & 26.8 & 44.0 \\
\multicolumn{3}{l|}{\textit{Ours (Augmentations)}}      & RN-101 & 81.0 & 33.4 & 84.3 & 32.9 & 27.6 & 25.7 & 38.3 & 47.0 & 86.5 & 36.9 & 84.9 & 64.6 & 28.7 & 85.8 & 42.3 & 40.2 & 1.5 & 33.7 & 41.8 & 48.3 \\
\arrayrulecolor{gray!30}
\midrule
\arrayrulecolor{black}
\multicolumn{3}{l|}{\textit{Ours (Augmentations) + MS}} & RN-101 & 91.6 & 51.2 & 84.7 & 37.3 & 29.1 & 24.6 & 31.3 & 37.2 & 86.5 & 44.3 & 85.3 & 62.8 & 22.6 & 87.6 & 38.9 & 52.3 & 0.65 & 37.2 & 50.0 & \textbf{50.3} \\

\bottomrule
\end{tabular}
\vspace{2mm}
\caption{GTA5-to-Cityscapes results. RN-101 and WRN-38 refer to ResNet-101 and Wider-ResNet-38 architectures. }
\label{table:gta}
\end{center}
\end{table*}

\begin{table*}[h]
\fontsize{8}{9}\selectfont
\setlength{\tabcolsep}{0.30em}
\def\arraystretch{1.15}
\arrayrulecolor{gray}
\begin{center}
\addtolength{\leftskip} {-1.3cm} 
\addtolength{\rightskip}{-1.8cm}
\textbf{SYNTHIA-to-Cityscapes}

\vspace{1mm}
\begin{tabular}{lll|c|cccccccccccccccc|c|c}
    & Year & Method & Backbone & \rotatebox{90}{road} & \rotatebox{90}{sidewalk} & \rotatebox{90}{building} & \rotatebox{90}{wall$^{*}$} & \rotatebox{90}{fence$^{*}$} & \rotatebox{90}{pole$^{*}$} & \rotatebox{90}{light} & \rotatebox{90}{sign} & \rotatebox{90}{veg.} & \rotatebox{90}{sky} & \rotatebox{90}{person} & \rotatebox{90}{rider} & \rotatebox{90}{car} & \rotatebox{90}{bus} & \rotatebox{90}{motor} & \rotatebox{90}{bike} & mIoU-16 & mIoU-13 \\ 
\toprule
\multicolumn{22}{c}{No Adaptation} \\ 
\midrule 
    & & Source only	& RN-101 &	    17.7 & 15.0 & 74.3 & 10.1 & 0.1 & 25.5 & 6.3 & 10.2 & 75.5 & 77.9 & 57.1 & 19.2 & 31.2 & 31.2 & 10.0 & 20.1 & 30.1 & 34.3\\ 
\midrule
\multicolumn{22}{c}{Adversarial Methods} \\ 
\midrule 
\cite{adaptsegnet}                  & \textit{2018} \hspace{2mm} &  AdaptSegNet      & RN-101  &	 79.2 &  37.2 &  78.8 &  10.5 &  0.3  &  25.1 &  9.9 &  10.5 &  78.2 &  80.5 &  53.5 &  19.6 &  67.0 &  29.5 &  21.6 &  31.3 &  39.5 &  45.9 \\ 
\cite{adaptsegnet}                  & \textit{2018} \hspace{2mm} &  AdaptSegNet-LS	             & RN-101  &	 84.0 &  40.5 &  79.3 &  10.4 &  0.2  &  22.7 &  6.5 &  8.0  &  78.3 &  82.7 &  56.3 &  22.4 &  74.0 &  33.2 &  18.9 &  34.9 &  40.8 &  47.6 \\ 
\cite{advent}                       & \textit{2019} \hspace{2mm} &  AdvEnt                 &  RN-101 &  87.0 &  44.1 &  79.7 &  9.6  &  0.6 &  24.3 &  4.8  &  7.2  &  80.1 &  83.6 &  56.4 &  23.7 &  72.7 &  32.6 &  12.8 &  33.7   &  40.8 &  47.6 \\
\cite{advent}                       & \textit{2019} \hspace{2mm} &  AdvEnt+MinEnt          &  RN-101 &  85.6 &  42.2 &  79.7 &  8.7  &  0.4 &  25.9 &  5.4  &  8.1  &  80.4 &  84.1 &  57.9 &  23.8 &  73.3 &  36.4 &  14.2 &  33.0   &  41.2 &  48.0 \\
\cite{discpatch}                    & \textit{2019} \hspace{2mm} &  Patch-Disc             &  RN-101 &  82.4 &  38.0 &  78.6 &  8.7  &  0.6 &  26.0 &  3.9  &  11.1 &  75.5 &  84.6 &  53.5 &  21.6 &  71.4 &  32.6 &  19.3 &  31.7   &  40.0 &  46.5 \\
\cite{zhang2019category}            & \textit{2019} \hspace{2mm} &  CAG                    &  RN-101 &  84.7 &  40.8 &  81.7 &  7.8  &  0.0 &  35.1 &  13.3 &  22.7 &  84.5 &  77.6 &  64.2 &  27.8 &  80.9 &  19.7 &  22.7 &  48.3   &  44.5 &  52.6 \\
\cite{Wang2020ClassesMA}            & \textit{2020} \hspace{2mm} &  FADA                   &  RN-101 &  84.5 &  40.1 &  83.1 &  4.8  &  0.0 &  34.3 &  20.1 &  27.2 &  84.8 &  84.0 &  53.5 &  22.6 &  85.4 &  43.7 &  26.8 &  27.8   &  45.2 &  52.5 \\
\cite{huang2020contextual}          & \textit{2020} \hspace{2mm} &  CrCDA R                &  RN-101 &  86.2 &  44.9 &  79.5 &  8.3  &  0.7 &  27.8 &  9.4  &  11.8 &  78.6 &  86.5 &  57.2 &  26.1 &  76.8 &  39.9 &  21.5 &  32.1   &  42.9 &  50.0 \\
\midrule 
\multicolumn{22}{c}{Image-to-Image / Style Transfer Methods} \\ 
\midrule 
\cite{li2019bidirectional}         & \textit{2019} \hspace{2mm} &  BDL (M2-F2)        &  RN-101 &  86.0 &  46.7 &  80.3 &  -   &  -   &  -    &  14.1 &  11.6 &  79.2 &  81.3 &  54.1 &  27.9 &  73.7 &  42.2 &  25.7 &  45.3 &  -    &  51.4 \\
\cite{chen2019crdoco}              & \textit{2019} \hspace{2mm} &  CrDoCo             &  DRN-26 &  62.2 &  21.2 &  72.8 &  4.2 &  0.8 &  30.1 &  4.1  &  10.7 &  76.3 &  73.6 &  45.6 &  14.9 &  69.2 &  14.1 &  12.2 &  23.0 &  33.4 &  38.5 \\
\cite{yang2020fda}                 & \textit{2020} \hspace{2mm} &  FDA (Ensemble)     &  RN-101 &  79.3 &  35.0 &  73.2 &  -   &  -   &  -    &  19.9 &  24.0 &  61.7 &  82.6 &  61.4 &  31.1 &  83.9 &  40.8 &  38.4 &  51.1 &  -    &  52.5 \\
\cite{kim2020learning}             & \textit{2020} \hspace{2mm} &  LTIR               &  RN-101 &  92.6 &  53.2 &  79.2 &  -   &  -   &  -    &  1.6  &  7.5  &  78.6 &  84.4 &  52.6 &  20.0 &  82.1 &  34.8 &  14.6 &  39.4 &  -    &  49.3 \\
\midrule 
\multicolumn{22}{c}{Self-Training Methods} \\ 
\midrule 
\cite{cbst}                        & \textit{2018} \hspace{2mm} &  CBST-SP            &  WRN-38 &  53.6 &  23.7 &  75.0 &  12.5 &  0.3  &  36.4 &  23.5 &  26.3 &  84.8 &  74.7 &  67.2 &  17.5 &  84.5 &  28.4 &  15.2 &  55.8 &  42.5    &  48.4\\
\cite{advent}                      & \textit{2019} \hspace{2mm} &  MinEnt             &  RN-101 &  73.5 &  29.2 &  77.1 &  7.7  &  0.2  &  27.0 &  7.1  &  11.4 &  76.7 &  82.1 &  57.2 &  21.3 &  69.4 &  29.2 &  12.9 &  27.9 &  38.1    &  44.2\\
\cite{maxsquare}                   & \textit{2019} \hspace{2mm} &  MaxSquare (MS)     &  RN-101 &  77.4 &  34.0 &  78.7 &  5.6  &  0.2  &  27.7 &  5.8  &  9.8  &  80.7 &  83.2 &  58.5 &  20.5 &  74.1 &  32.1 &  11.0 &  29.9 &  39.3    &  45.8\\
\cite{maxsquare}                   & \textit{2019} \hspace{2mm} &  MS + IW + Multi &  RN-101 &  82.9 &  40.7 &  80.3 &  10.2 &  0.8  &  25.8 &  12.8 &  18.2 &  82.5 &  82.2 &  53.1 &  18.0 &  79.0 &  31.4 &  10.4 &  35.6 &  41.4    &  48.2\\
\cite{zou2019confidence}           & \textit{2019} \hspace{2mm} &  CRST (MRENT)       &  RN-101 &  69.6 &  32.6 &  75.8 &  12.2 &  1.8  &  35.3 &  23.3 &  29.5 &  77.7 &  78.9 &  60.0 &  28.5 &  81.5 &  25.9 &  19.6 &  41.8 &  43.4    &  49.6 \\
\cite{zou2019confidence}           & \textit{2019} \hspace{2mm} &  CRST (MRKLD)       &  RN-101 &  67.7 &  32.2 &  73.9 &  10.7 &  1.6  &  37.4 &  22.2 &  31.2 &  80.8 &  80.5 &  60.8 &  29.1 &  82.8 &  25.0 &  19.4 &  45.3 &  43.8    &  50.1 \\
\cite{zou2019confidence}           & \textit{2019} \hspace{2mm} &  CRST (LRENT)       &  RN-101 &  65.6 &  30.3 &  74.6 &  13.8 &  1.5  &  35.8 &  23.1 &  29.1 &  77.0 &  77.5 &  60.1 &  28.5 &  82.2 &  22.6 &  20.1 &  41.9 &  42.7    &  48.7 \\
\cite{lian2019constructing}        & \textit{2019} \hspace{2mm} & PyCDA &  RN-101             &  75.5   &  30.9 &  83.3 &  20.8 &  0.7  &  32.7 &  27.3 &  33.5 &  84.7 &  85.0 &  64.1 &  25.4 &  85.0 &  45.2 &  21.2 &  32.0 &  46.7 &  53.3 \\ 
\cite{iqbal2020mlsl}               & \textit{2020} \hspace{2mm} &  MLSL (SISC)        &  RN-101 &  73.7 &  34.4 &  78.7 &  13.7 &  2.9  &  36.6 &  28.2 &  22.3 &  86.1 &  76.8 &  65.3 &  20.5 &  81.7 &  31.4 &  13.9 &  47.3 &  44.4    &  50.8 \\
\cite{iqbal2020mlsl}               & \textit{2020} \hspace{2mm} &  MLSL (+PWL)    &  RN-101 &  59.2 &  30.2 &  68.5 &  22.9 &  1.0  &  36.2 &  32.7 &  28.3 &  86.2 &  75.4 &  68.6 &  27.7 &  82.7 &  26.3 &  24.3 &  52.7 &  45.2    &  51.0 \\
\midrule 
\multicolumn{22}{c}{Other Methods} \\ 
\midrule 
\cite{lee2019sliced}               & \textit{2019} \hspace{2mm} &  SWD     &  RN-101 &  82.4 &  33.2 &  82.5 &  -    &  -   &  -    &  22.6 &  19.7 &  83.7 &  78.8 &  44.0 &  17.9 &  75.4 &  30.2  &  14.4 &  39.9 &  -    &  48.1 \\
\cite{li2020content}               & \textit{2020} \hspace{2mm} &  CCM     &  RN-101 &  79.6 &  36.4 &  80.6 &  13.3 &  0.3 &  25.5 &  22.4 &  14.9 &  81.8 &  77.4 &  56.8 &  25.9 &  80.7 &  45.27 &  29.9 &  52.0 &  45.2 &  52.9 \\
\cite{kang2020pixel}               & \textit{2020} \hspace{2mm} &  PLCA    &  RN-101 &  82.6 &  29.0 &  81.0 &  11.2 &  0.2 &  33.6 &  24.9 &  18.3 &  82.8 &  82.3 &  62.1 &  26.5 &  85.6 &  48.9  &  26.8 &  52.2 &  \textbf{46.8} &  54.0 \\
\cite{pan2020unsupervised}         & \textit{2020} \hspace{2mm} &  IntraDA &  RN-101 &  84.3 &  37.7 &  79.5 &  5.3  &  0.4 &  24.9 &  9.2  &  8.4  &  80.0 &  84.1 &  57.2 &  23.0 &  78.0 &  38.1  &  20.3 &  36.5 &  41.7 &  48.9 \\
\midrule \midrule 
& &  \textit{Ours (CutMix)}       &  RN-101 &  84.8 &  42.1 &  81.0 &  6.98 &  0.25 &  27.9 &  15.6 &  16.6 &  82.3 &  80.7 &  53.9 &  21.8 &  83.1 &  39.3 &  21.0 &  43.1 &  43.8 &  51.2 \\
& & \textit{Ours (Augmentations)} &  RN-101 &  73.0 &  38.9 &  70.8 &  6.0  &  0.1  &  27.0 &  17.0 &  20.3 &  83.0 &  84.2 &  59.1 &  27.0 &  80.1 &  37.4 &  17.8 &  52.4 &  43.4 &  50.8 \\
& & \textit{Ours (Fourier) }      &  RN-101 &  82.9 &  36.2 &  81.7 &  9.74 &  0.11 &  29.7 &  16.7 &  19.2 &  84.3 &  84.2 &  62.3 &  16.9 &  84.6 &  39.4 &  3.04 &  52.5 &  44.0 &  51.1 \\
& & \textit{Ours (Style)}         &  RN-101 &  82.1 &  38.0 &  76.2 &  3.97 &  0.12 &  26.4 &  14.2 &  11.0 &  75.5 &  70.7 &  54.3 &  20.6 &  75.8 &  36.9 &  19.4 &  39.3 &  40.3 &  47.3 \\
\arrayrulecolor{gray!30}
\midrule
\arrayrulecolor{black}
& & \textit{Ours (Fourier + CutMix)}         &  RN-101 &  92.5 & 54.6 & 79.8 & 4.78 & 0.08 & 24.1 & 22.8 & 17.8 & 79.4 & 76.5 & 60.8 & 24.7 & 85.7 & 33.5 & 26.4 & 54.4 & 46.1 & \textbf{54.5} \\
 \bottomrule
\end{tabular}
\vspace{2mm}
\caption{Results on the SYNTHIA-to-Cityscapes benchmark. mIoU-16 and mIoU-13 refer to mean intersection-over-union on the standard sets of 16 and 13 classes, respectively.}
\label{table:synthia}
\end{center}  
\end{table*}


\begin{figure*}[!ht]
  \centering
  \makebox[\textwidth][c]{\includegraphics[width=1.0\textwidth]{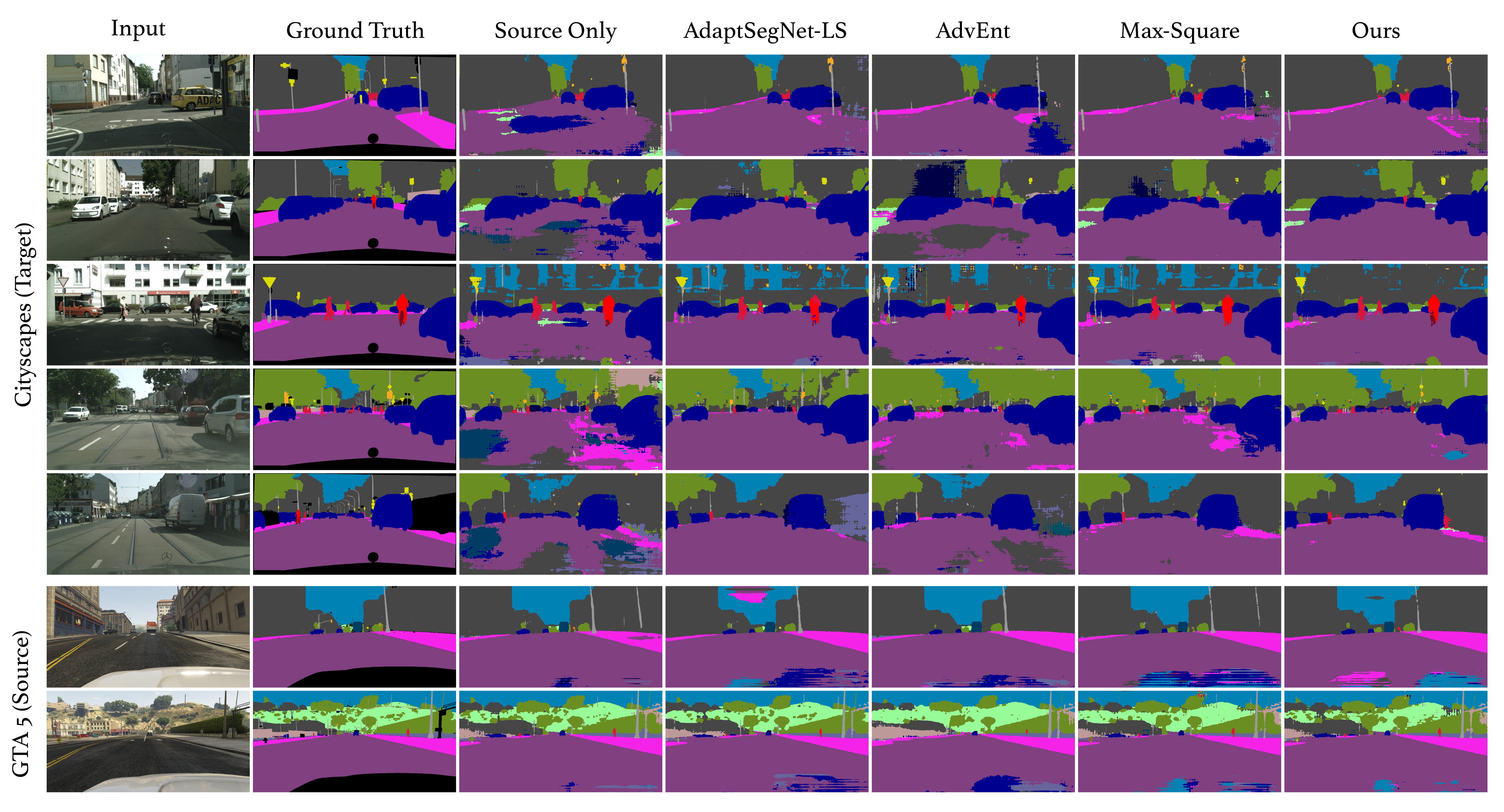}}%
  \caption{Qualitative examples of our consistency training method and prior methods on GTA5-to-Cityscapes. The column "Ours" refers to our baseline model with augmentation-based perturbations. Note that these images are not hand-picked; they are the first 5 images in the Cityscapes validation set. Visually, our method yields better segmentations than prior techniques. On the third image, for example, it is the only method that properly segments the bicycle (shown in red) and the only method that does not mistake a large proportion of the building for sky. }
  \label{fig:qualitative_gta}
\end{figure*}

\paragraph{Results}

We present our results on the highly competitive GTA5-to-Cityscapes and SYNTHIA-to-Cityscapes benchmarks in Tables \ref{table:gta} and \ref{table:synthia}. For comparison, we have included 27 other methods from the past two years, including all top results from the SoTa benchmark list on \href{https://paperswithcode.com/sota/synthetic-to-real-translation-on-gtav-to}{PapersWithCode}. Our simple consistency training method performs extremely well relative to other methods. 

\begin{figure*}[!h]
  \centering
  \makebox[\textwidth][c]{\includegraphics[width=1.2\textwidth]{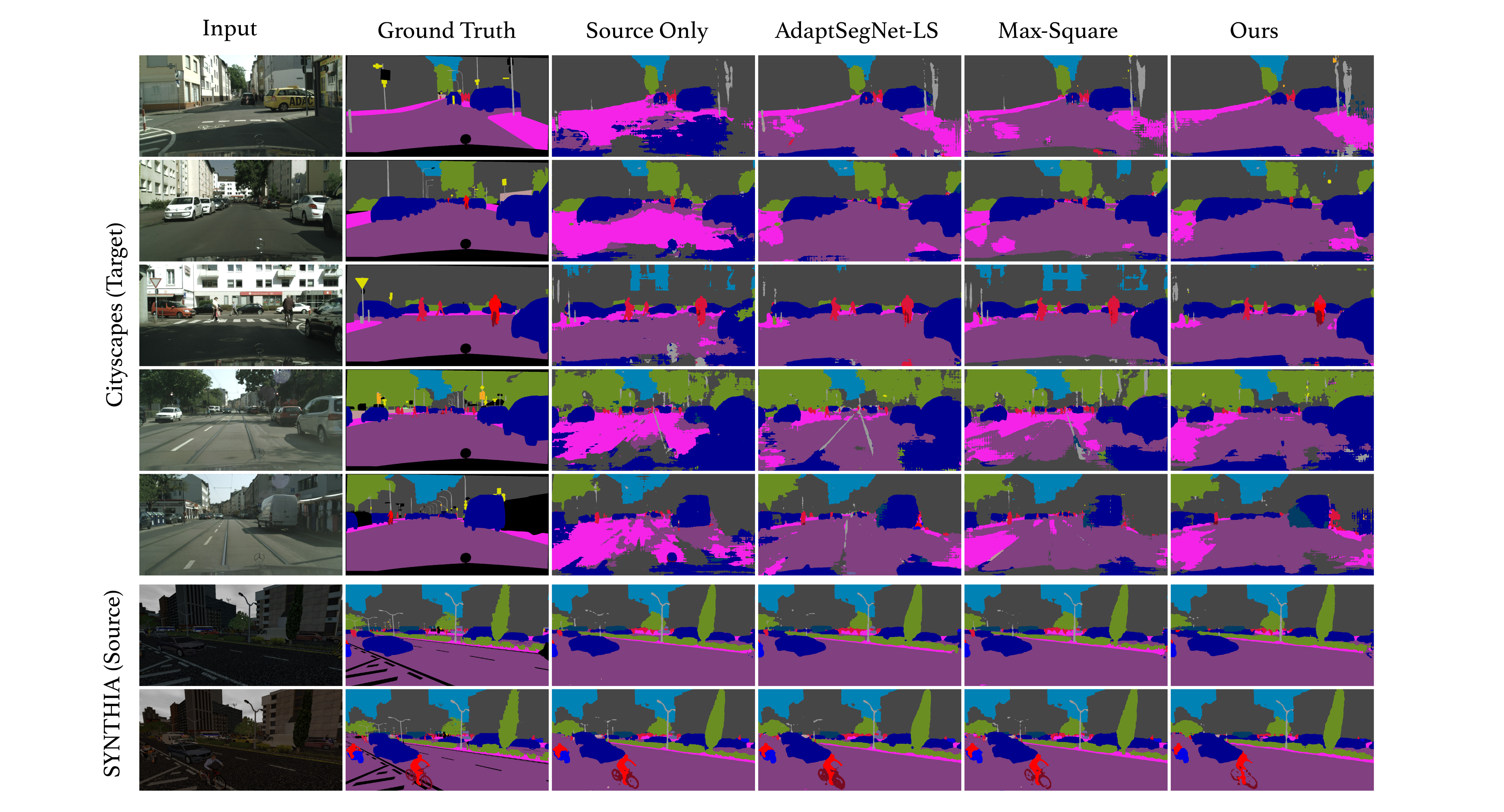}}%
  \caption{Qualitative examples of our consistency training method and prior methods on SYNTHIA-to-Cityscapes.}
  \label{fig:qualitative_synthia}
\end{figure*}

We also show qualitative results in Figures \ref{fig:qualitative_gta} and \ref{fig:qualitative_synthia}.

\section{Analysis}

\paragraph{Overall} 

On GTA5-to-Cityscapes (\autoref{table:gta}), our baseline augmentation-based model performs best, followed by CutMix-, Fourier-, and finally style-based consistency. When combined with entropy minimization (in the form of the max-square loss), our baseline method attains an mIoU of 50.3, outperforming all other methods with the exception of a single method using a distillation of multiple models that performs 0.1 mIoU better (50.4). 

On SYNTHIA-to-Cityscapes (\autoref{table:synthia}), CutMix-based training method performed best, followed by Fourier-, augmentation-, and again style-based consistency. Apart from style-based consistency, all are within 0.4 mIoU; they perform very competitively with more complex domain adaptation approaches. 

\paragraph{Fourier} It is natural to compare the results of our Fourier-based model to those from FDA \cite{yang2020fda}, which modifies the source images directly. On GTA5-to-Cityscapes, our method outperforms FDA (45.4 vs 44.6 mIoU). On SYNTHIA-to-Cityscapes, \cite{yang2020fda} only presents results for a distillation of an ensemble of models, not a single model, but nonetheless our single model only slightly underperforms the distilled ensemble (51.1 vs 52.5 mIoU). 

\vspace{-1mm}
\paragraph{CutMix} The CutMix-based model, which has no domain-adaptation-specific components, gives very strong results. This is notable because most highly-competitive recent methods employ domain-adaptation-specific components in the form of adversarial training (e.g. \cite{Wang2020ClassesMA}), domain-specific priors (e.g. \cite{iqbal2020mlsl}), style transfer (e.g. \cite{kim2020learning}), or distribution alignment (e.g. \cite{lee2019sliced}). Compared to other pure self-training methods (e.g. MinEnt \cite{advent}, MaxSquare \cite{maxsquare}, CRST \cite{zou2019confidence}, PyCDA \cite{lian2019constructing}) our method performs substantially better. 

\vspace{-1mm}
\paragraph{Style} The style-based variant performs badly relative to our other models and to other approaches that use style transfer directly. We hypothesize that this underperformance may be due to our choice of style transfer model, as we used AvatarNet \cite{sheng2018avatar} rather than CycleGAN as in \cite{li2019bidirectional} or CycleGAN + StyleSwap as in \cite{kim2020learning}.

\paragraph{Combining Perturbation Functions}

PixMatch also makes it easy to combine multiple perturbation functions. 
For example, \autoref{table:synthia} shows the result of combining the Fourier- and CutMix-based perturbations on the SYNTHIA benchmark. We give additional results for combining perturbation functions in the Appendix (\autoref{table:new_experiments}). We find that combining two perturbation functions often improves results, while combining three or more is ``too much'' perturbation. 

We note that we have not experimented with using different strengths/weights for different perturbation functions, nor have we experimented with stochastically applying perturbation functions. Studying the interactions of multiple perturbation functions could be a promising avenue for future research. 

\section{Ablation Studies and Further Experiments}

Below, we present an extensive set of ablation studies on our baseline (augmentation-based) model. 

\paragraph{Ablation: Varying $\lam_T$}

Our proposed consistency training method involves choosing a perturbation function and tuning the single hyperparameter $\lam_T$, which trades off between the strength of the source supervised loss and the target consistency loss. We present results for $\lam_T = 0.00, 0.05, 0.10, 0.15,$ and $0.20$ in Table \ref{table:ablation}, finding that $\lam_T = 0.10$ and $0.15$ both yield strong results. This finding shows a stark contrast between UDA for semantic segmentation and self-supervised learning for image classification, where the consistency loss is usually much greater than 1 ($\lambda_{unsup} > 1$) \cite{fixmatch,xie2019unsupervised}.


\begin{table}[t]
  \def\arraystretch{1.4}
  \arrayrulecolor{gray}
  \begin{center}
  \vspace{2mm}
  \begin{tabular}{lccccc} \hline
  $\lam_T \qquad$ & \textit{0.00 (Source)}   & \textit{0.05} & \textit{0.10} & \textit{0.15} & \textit{0.20} \\ \toprule 
  \textit{mIoU} & 36.9 & 48.7 & 49.4 & \textbf{49.8} & 48.0 \\ \bottomrule
  \end{tabular}
  \vspace{2mm}
  \caption{Results of our augmentation-based model, varying the consistency loss weight hyperparameter $\lam_T$.}
  \label{table:ablation}
  \end{center}
\end{table}


\begin{table}[t]
  \def\arraystretch{1.4}
  \arrayrulecolor{gray}
  \begin{center}
  \vspace{2mm}
  \begin{tabular}{lcccc} \hline
  $\tau \qquad$ & \textit{0.00}   & \textit{0.50} & \textit{0.90} & \textit{0.95} \\ \toprule 
  \textit{mIoU} & \textbf{49.9} & 48.7 & 47.1 & 47.0 \\ \bottomrule
  \end{tabular}
  \vspace{2mm}
  \end{center}
  \caption{Results of our augmentation-based model with an additional pseudolabel threshold. A threshold of $\tau \in [0,1)$ corresponds to only calculating the consistency loss on pseudolabeled pixels for which the model assigns an output probability greater than $\tau$ to one class.}
  \label{table:threshold}
\end{table}

\begin{table}[t]
  \def\arraystretch{1.4}
  \arrayrulecolor{gray}
  \begin{center}
  \vspace{2mm}
  \begin{tabular}{lcccc} \hline
  $\lam_{MSL} \qquad$ & \textit{0.00}   & \textit{0.05} & \textit{0.10} & \textit{0.15} \\ \toprule \midrule 
  \textit{mIoU} & 49.9 & \textbf{50.7} & 50.5 & 50.4 \\ \bottomrule
  \end{tabular}
  \vspace{2mm}
  \caption{Results of a model combining the max-square loss \cite{maxsquare} with our augmentation-based consistency loss, for different values of $\lam_{MSL}$}
  \label{table:combining}
  \end{center}
\end{table}

\paragraph{Ablation: Adding a Pseudolabel Threshold} 

It is also possible to add a pseudolabel threshold in our formulation of consistency training. This corresponds to only pseudolabeling pixels in which the target output probability exceeds some threshold $\tau$ (other pixels are ignored from the loss). For example, \cite{fixmatch} uses $\tau = 0.95$. 

In Table \ref{table:threshold}, we show results for $\tau = 0, 0.5, 0.9,$ and $0.95$. Surprisingly, unlike in semi-supervised learning for image classification, we find that $\tau = 0.0$ (i.e. no thresholding) performs best. We hypothesize that this may be because when a large threshold is used, the loss is dominated by easy classes; if this is the case, future work could potentially address the issue by performing consistency training with class-wise thresholds \cite{cbst}.

\paragraph{Combining PixMatch with Other Methods}

Our consistency training method is complementary to a wide range of other domain adaptation techniques. To demonstrate this, we combine our method with the max-square loss proposed in \cite{maxsquare}. That is, we modify our loss function to be $\mathcal{L} = \mathcal{L}_S + \lambda_T \mathcal{L}_T + \lambda_{MSL} \mathcal{L}_{MSL}$. In Table \ref{table:combining}, we fix $\lam_T = 0.10$ and present results for different weightings $\lam_{MSL}$ of the max-square loss term. The across-the-board performance improvement demonstrates that our method is readily combinable with other self-training approaches. As a highlight, the variant with $\lam_{MSL} = 0.05$ achieves state-of-the-art performance on GTA5-to-Cityscapes.

\section{Conclusion}

We present PixMatch, an unsupervised domain adaptation approach for semantic segmentation that incorporates consistency training in the target domain. Through an extensive set of ablation studies, we sought to understand which aspects of this consistency training framework are most important to the model's final performance. In comparison to adversarial approaches that have dominated the recent literature, our approach is faster, more stable, has superior performance, and may be readily combined with other methods. Future work may explore new perturbation functions, the combination of multiple perturbation functions, or the application of this framework to other tasks (e.g. object detection). We will release code and pretrained models to facilitate future research.

\clearpage 

\nocite{*}
{\small
\bibliographystyle{ieee_fullname}
\bibliography{egbib}
}

\clearpage 

\section*{Appendix}  

\subsection{Perturbation Function Details}

Here, we describe the parameters used for each perturbation function. For the baseline model with data augmentation as the perturbation function, we use the set of augmentation in Figure \ref{fig:sup_algos}. These are based on the augmentations from SimCLR \cite{simclr} and implemented using the albumentations \cite{albumentations} library. For style consistency model, we obtain perturbed images by applying the Avatar-Net \cite{sheng2018avatar} style transfer model to the image.\footnote{\url{https://github.com/tyui592/Avatar-Net_Pytorch}} We use the default Avatar-Net strength parameter of 0.1. For the Fourier consistency model, we employ the same Fourier swapping function as \cite{yang2020fda}. We use a low-frequency window size parameter of $\beta=0.01$. Finally, for the CutMix consistency model, we employ the cutmix function from \cite{yun2019cutmix}. \footnote{\url{https://github.com/clovaai/CutMix-PyTorch}}

\subsection{Additional Combinations of Perturbation Functions}

In \autoref{table:new_experiments}, we show the result of additional combinations of perturbation functions. 

\newpage

\begin{algorithm}[t!]
  \fontsize{8}{9}\fontfamily{pcr}\selectfont
  \SetAlgoLined
  \DontPrintSemicolon
  \hskip0.1em Compose([ \;
  \hskip1.1em  RandomResizedCrop(scale=(0.2, 1)),\;
  \hskip1.1em  Compose([ \;
  \hskip2.1em    RandomBrightnessContrast(p=1), \;
  \hskip2.1em    HueSaturationValue(p=1) \;
  \hskip1.1em  ], p=0.8),  \;
  \hskip1.1em  ToGray(p=0.2), \;
  \hskip1.1em  GaussianBlur(blur\_limit=5, p=0.5),\;
  \hskip0.1em]) \;
  \caption{\small Augmentations}
\end{algorithm}
\begin{figure}[h!]
  \caption{PyTorch-style pseudocode for our chosen data augmentation function.}
  \label{fig:sup_algos}
\end{figure}

\clearpage

\begin{figure*}
  \centering
  \includegraphics[width=\textwidth]{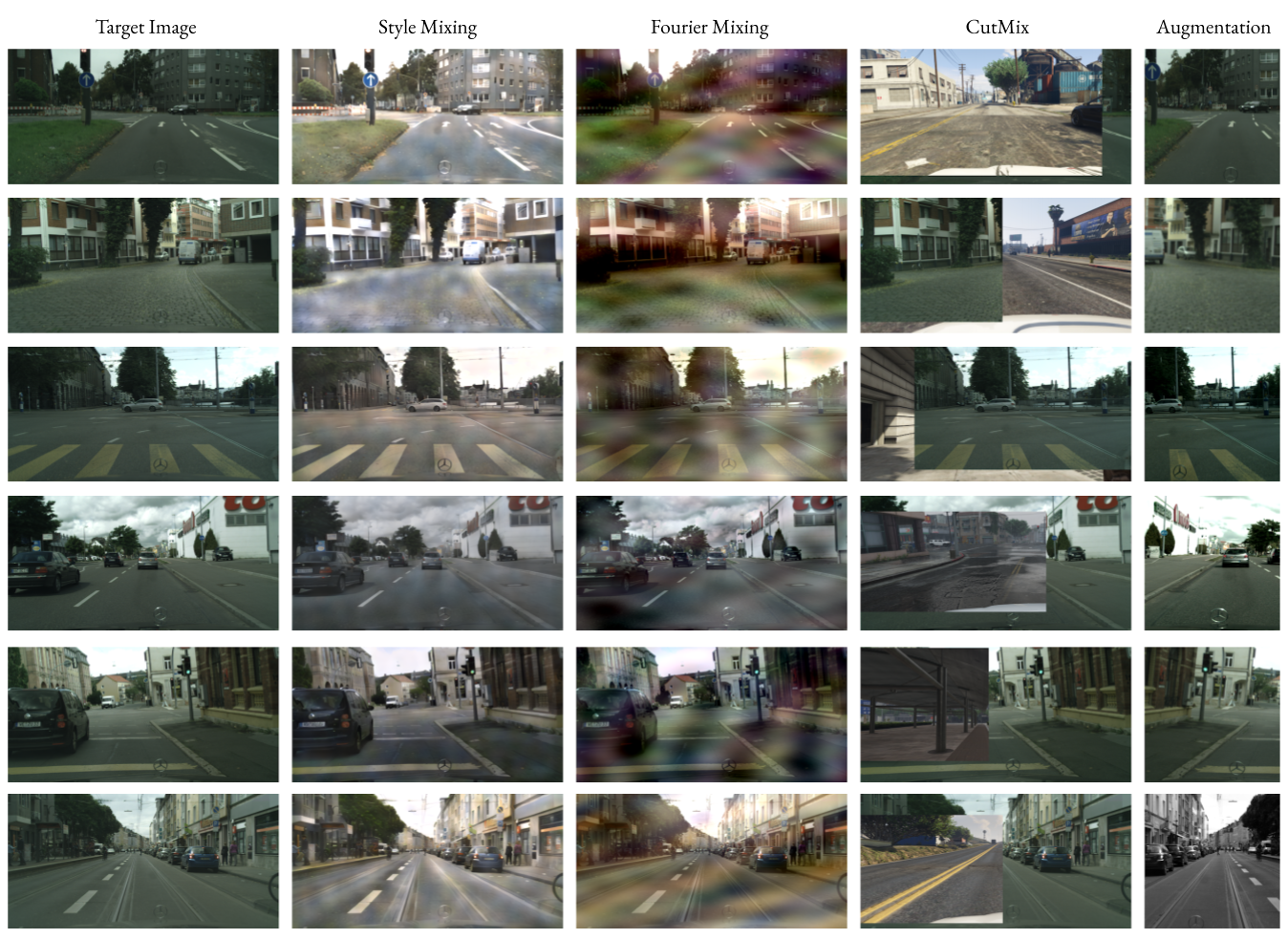}
  \caption{Examples of our data augmentation function applied to target-domain images. Note that the ``augmentation'' images are a different size and aspect ratio because random cropping is part of the augmentation procedure. Also note that although the ``style'' and ``Fourier'' images appear unnatural on a global scale, they appear similar to the source domain on a local/low-frequency scale.}
  \label{fig:sup_vis}
\end{figure*}

\clearpage

\clearpage

\begin{table*}
\begin{center}

\textbf{SYNTHIA-to-Cityscapes}

\vspace{1mm}
\begin{tabular}{rl|c|c}
&    & mIoU-16 & mIoU-13 \\ \hline \addlinespace[1mm]
& LTIR [28]                        & -       & 49.3    \\
& SOTA (PLCA [27])                 & \textbf{46.8}    & 54.0    \\ \addlinespace[1mm] \hdashline \addlinespace[1mm]
& Ours: Fourier only                     & 44.0    & 51.1    \\
& Ours: Fourier + Aug                    & 45.6    & 54.0    \\
& Ours: Fourier + Cutmix                 & 46.1    & \textbf{54.5}    \\
& Ours: Fourier + Cutmix + Aug           & 43.7    & 52.1   
\end{tabular}

\vspace{5mm}

\textbf{GTA-to-Cityscapes}

\vspace{1mm}
\begin{tabular}{rl|cc}
  &    & mIoU           & \hspace{8mm}  \\ \hline \addlinespace[1mm]
              & SOTA (LTIR [28])                 & 50.2           &   \\
              & PLCA [27]                        & 47.7           &   \\ \hdashline \addlinespace[1mm]
              & Ours: Aug only                         & 48.3           &   \\
              & Ours: Aug + Fourier                    & 49.3           &   \\
              & Ours: Aug + MaxSquareLoss \hspace{8mm}              & \textbf{50.4}  & 
\end{tabular}
\end{center}
\vspace{-2mm}
\caption{Performance of additional combinations of perturbation functions compared to SOTA.}
\vspace{-5mm}
\label{table:new_experiments}
\end{table*}

\end{document}